\documentclass[twoside]{article}

\usepackage{aistats2020}
\usepackage[round]{natbib}

\usepackage{amsmath,amssymb,amsthm}
\usepackage{algorithm,algpseudocode}
\usepackage{graphicx}
\usepackage{multirow}
\usepackage{booktabs}

\newtheorem{theorem}{Theorem}[section]
\newtheorem{lemma}[theorem]{Lemma}
\newtheorem{corollary}[theorem]{Corollary}
\newtheorem{definition}[theorem]{Definition}
\newtheorem{assumption}[theorem]{Assumption}

\newcommand\independent{\protect\mathpalette{\protect\independenT}{\perp}}
\def\independenT#1#2{\mathrel{\rlap{$#1#2$}\mkern2mu{#1#2}}}
\newcommand{\E}{\mathbb{E}}
\renewcommand{\P}{\mathbb{P}}
\newcommand{\X}{\mathcal{X}}
\newcommand{\Y}{\mathcal{Y}}

\newcommand{\cR}{\mathcal{R}}

\newcommand{\IPM}{\mathrm{IPM}}

\newcommand{\R}{\mathbb{R}}

\newcommand{\hatepsPehe}{{\hat{\epsilon}_\mathrm{PEHE}}}
\newcommand{\epsPehe}{{\epsilon_\mathrm{PEHE}}}
\newcommand{\epsRct}{{\epsilon_\mathrm{RCT}}}
\newcommand{\epsFf}{{\epsilon(f,f^*)}}
\newcommand{\epsF}{{\epsilon_{\mathrm{F}}}}
\newcommand{\epsCF}{{\epsilon_{\mathrm{CF}}}}

\newcommand{\A}{\mathcal{A}}
\newcommand{\Ra}{\mathcal{R}}
\newcommand{\C}{\mathcal{C}}

\newcommand{\M}{\mathcal{M}}
\newcommand{\DD}{\mathcal{D}}
\newcommand{\D}{D}

\newcommand{\DF}{D_\mathrm{F}}

\newcommand{\T}{\mathcal{T}}

\begin{document}

%

%

\twocolumn[

\aistatstitle{Learning Interpretable Models with Causal Guarantees}

\aistatsauthor{ Carolyn Kim \And Osbert Bastani }

\aistatsaddress{ Stanford University \And University of Pennsylvania } ]

\begin{abstract}
Machine learning has shown much promise in helping improve the quality of medical, legal, and financial decision-making. In these applications, machine learning models must satisfy two important criteria: (i) they must be causal, since the goal is typically to predict individual treatment effects, and (ii) they must be interpretable, so that human decision makers can validate and trust the model predictions. There has recently been much progress along each direction independently, yet the state-of-the-art approaches are fundamentally incompatible. We propose a framework for learning interpretable models from observational data that can be used to predict individual treatment effects (ITEs). In particular, our framework converts \emph{any} supervised learning algorithm into an algorithm for estimating ITEs. Furthermore, we prove an error bound on the treatment effects predicted by our model. Finally, in an experiment on real-world data, we show that the models trained using our framework significantly outperform a number of baselines.

\end{abstract}

\section{Introduction}

Machine learning is increasingly being used to help inform consequential decisions in healthcare, law, and finance. The goal is often to predict the effect of an intervention on an individual (called an \emph{individual treatment effect})---e.g., the efficacy of a drug on a patient~\citep{international2009estimation,kim2011battle,bastani2015online,henry2015targeted}, whether a defendent in a court case is a flight risk~\citep{kleinberg2017human}, or whether an applicant will repay a loan~\citep{hardt2016equality}.

There are two important properties that these machine learning models must satisfy: (i) they must be must be causal~\citep{potentialOutcomes,pearl2010causal}, and (ii) they must be interpretable. First, to predict treatment effects, our model must predict outcomes when the world is modified in some way (called a \emph{counterfactual outcome}). For example, to predict the efficacy of a drug on a patient, we need to know the patient's outcome both when given the drug and when not given the drug. One way to predict counterfactual outcomes is to use randomized controlled experiments (RCTs)---by randomly assigning individuals to treatment and control groups, we can ensure that the model generalizes to predicting counterfactual outcomes. However, RCT data is often too expensive to obtain.\footnote{ITEs are also known as heterogeneous treatment effects or conditional average treatment effects.}
Instead, many approaches consider predicting ITEs using \emph{observational data}, where individuals are selected into treatment and control groups by unknown mechanisms~\citep{potentialOutcomes,shalit2017}---for example, honest trees~\citep{athey2016recursive}, causal forests~\citep{wager2017estimation}, propensity score weighting~\citep{austin2011introduction,shi2019adapting}, and causal representations~\citep{johansson2016learning,shalit2017}.

Second, the learned model must be interpretable---i.e., a human domain expert (e.g., a doctor) must be able to validate the model. Interpretability is important since there are often defects in the training data that cause the model to make preventable errors. Indeed, it has been shown that these issues often arise in practice, and that interpretability can help experts diagnose these issues~\citep{caruana2015intelligible,ribeiro2016model,bastani2017interpreting}. As a consequence, many algorithms have been proposed for learning interpretable models, including decision trees~\citep{breiman2017classification,bastani2017interpreting}, sparse linear models~\citep{tibshirani1996regression,ustun2016supersparse}, generalized additive models~\citep{lou2012intelligible,caruana2015intelligible}, rule lists~\citep{wang2015falling,yang2017scalable,angelino2017learning}, decision sets~\citep{lakkaraju2016interpretable}, and programs~\citep{ellis2015unsupervised,verma2018programmatically,valkov2018houdini,ellis2018learning}.

However, while there has been work on learning causal models and on learning interpretable models, there has been relatively little work on designing algorithms that are capable of achieving both desirable properties. One proposed approach is the ``honest tree'' algorithm for learning decision trees for prediting ITEs~\citep{athey2016recursive}. Outside of this approach, most state-of-the-art approaches largely leverage techniques that are specific to learning blackbox models---e.g., using neural networks to learn representations~\citep{wager2017estimation,shi2019adapting} or learning ensemble models~\citep{wager2017estimation}. These techniques often rely crucially on the blackbox nature of the model family, and cannot be adapted to learning interpretable models. For example, the causal representations approach relies on learning an intermediate representation $\Phi:\X\to\cR$~\citep{shalit2017}, and then using supervised learning to train a model $h:\cR\to\Y$. Even in the best case, $\Phi$ is linear and $h$ is interpretable (e.g., a decision tree), then the composition $f(x,t)=h(\Phi(x),t)$ is not interpretable (e.g., a decision tree where the internal branches are linear functions of $x\in\X$).

We propose a general framework for learning interpretable models for ITE prediction. Given (i) \emph{any} interpretable supervised learning algorithm $\A$, and (ii) a blackbox \emph{oracle model} $f^*$ for ITE prediction, it learns an interpretable model $\hat f$ for ITE prediction. It does so using model compression~\citep{bucilua2006model,hinton2015distilling}---i.e., it uses $\A$ to train $\hat f$ to approximate $f^*$ on a distribution $p(x, t)$, where $x\in\X$ are the covariates and $t\in\{0,1\}$ is the treatment indicator; then, we use
\begin{align*}
\hat{\tau}_f(x)=\hat{f}(x,1)-\hat{f}(x,0)
\end{align*}
to predict ITEs.

The key issue is choosing $p(x,t)$. We use the \emph{RCT distribution}, which is the distribution obtained by running an RCT---i.e., treatments $t$ are randomly assigned and are independent of the covariates $x$. Since RCTs can be used to predict ITEs, $\hat{f}$ should have good performance as long as $f^*$ has good performance and $\hat{f}$ is a good approximation of $f^*$ on the RCT distribution.

We prove theoretical guarantees on the performance of $\hat{f}$. We show that the performance of $\hat{f}$ breaks down into three parts: (i) the error of the oracle model $f^*$ on the RCT distribution, (ii) the error of the \emph{best} interpretable model $\tilde{f}$ on the RCT distribution, and (iii) the generalization error. Intuitively, terms (ii) and (iii) quantify the error we would get if we had access to data from the RCT distribution, and used $\mathcal{A}$ to train $\hat{f}$ using this data. Thus, our result can be interpreted as showing that the ``price'' of lacking access to RCT data is the error of the oracle model $f^*$ (in addition to a multiplicative constant). Second, we show that as a consequence, under the assumption of strong ignorability, we can use recent guarantees for oracle models based on the causal representations approach~\citep{johansson2016learning,shalit2017} to obtain end-to-end theoretical guarantees for $\hat{f}$.

Finally, we evaluate our approach and show how it can be used to improve the performance of a wide range of models. In particular, we consider a variety of supervised learning algorithms for different model families, and show that our algorithm improves performs across this entire range of algorithms and models. Our focus is on evaluating the improvement in the performance of our models. Because our framework is flexible and can be applied to \emph{any} supervised learning algorithm, the user can choose an interpretable machine learning algorithm that is most suitable for their application, and then leverage our framework to improve the performance of that algorithm.

We note that our approach has a number of important advantages compared to designing an algorithm that directly learns interpretable models for estimating ITEs. First, there are many interpretable learning algorithms for the supervised setting, and the choice of algorithm often depends on the problem domain. Adapting each of these approaches to estimating ITEs might be possible, but may require a new approach for each learning algorithm. Second, our approach can substantially outperform algorithms for directly learning interpretable models, since we can leverage sophisticated techniques such as causal representation learning that cannot be directly used to learn intepretable models. For example, in Section~\ref{sec:exp}, we show that empirically, our approach substantially outperforms honest trees~\citep{athey2016recursive}. Finally, as we describe in Section~\ref{sec:theory}, up to constant factors, we can recover convergence rates equal to those for supervised learning.

\textbf{Related work.}
The most closely related work is honest trees~\citep{athey2016recursive}. This work builds on CART~\citep{breiman2017classification}; they reduce the bias of CART by using different subsets of the training data to estimate the internal nodes and the leaf nodes. However, their approach is tailored to a specific interpretable model family (i.e., decision trees). Also, unlike their work, our approach comes with provable performance guarantees. Finally, we show in our experiments that our approach can substantially outperform theirs.

There has also been work using interpretability to identify causal issues in learned predictive models~\citep{caruana2015intelligible,ribeiro2016model,bastani2017interpreting}. However, there is currently no way to fix these causal issues except by having an expert manually correct the model. There has been a wide range of work using an uninterpretable oracle model $f^*$ to guide the learning of an interpretable model~\citep{lakkaraju2017interpretable,bastani2017interpreting,verma2018programmatically,frosst2017distilling,bastani2018verifiable}. Our work is the first to leverage this approach in the context of learning causal models.

Finally, there has been recent work on empirically evaluating the interpretability of different model families~\citep{doshi2017towards}. Since our framework can be applied to any interpretable supervised learning algorithm, a user can first use these approaches to choose a suitable interpretable supervised learning algorithm, and then use our framework to convert this algorithm to an algorithm for estimating ITEs. Furthermore, there has been work jointly optimizing interpretability and performance ~\citep{lage2018human}; we believe it is possible to integrate their approach with ours, but we leave this possibility to future work.

\section{Preliminaries}
\label{sec:back}

We use the Rubin-Neyman potential outcomes framework \citep{potentialOutcomes}. We are given a set of individuals (e.g., patients), and want to estimate the efficacy of a treatment (e.g., prescribing a drug) for each individual. Each individual is associated with covariates $X$ (e.g., healthcare history), and is assigned to either the control ($T=0$) or the treatment ($T=1$) group. Furthermore, each individual is associated with two \emph{potential outcomes} $Y_0$ if $T=0$ and $Y_1$ if $T=1$ (e.g., how fast the patient recovers). We want to estimate \emph{treatment effect} $Y_1-Y_0$, which indicates whether the outcome is better if treated (e.g., we should prescribe the drug if $Y_1-Y_0>0$). Formally, each individual is associated with a tuple of random variables $(X,T,Y_0,Y_1)$, where $X\in\X\subseteq\mathbb{R}^d$, the $T\in\T=\{0,1\}$, and $Y_0,Y_1\in\Y\subseteq\mathbb{R}$. We assume the tuple for each individual is drawn i.i.d. from $p(x,t,y_0,y_1)$.

The fundamental challenge in causal inference is that for each individual, we only observe either $Y_0$ or $Y_1$, but never both---in particular, for each individual, we only observe $(X,T,Y_T)$. The observed outcome $Y_T$ is the \emph{factual outcome}, and the unobserved outcome $Y_{1-T}$ is the \emph{counterfactual outcome}. For example, if we give a patient the drug, we cannnot observe what would have happened without the drug. Thus, we can only estimate the average $Y_1-Y_0$ over multiple individuals. If we average over the entire population, then we obtain \emph{average treatment effect (ATE)} $\text{ATE}=\E_p[Y_1-Y_0]$. However, the ATE does not yield any information about the efficacy of treatment on an individual. Instead, our goal is to estimate the efficacy of a treatment for an individual based on their covariates.
\begin{definition}
\rm
The \emph{individual treatment effect (ITE)} is
\begin{align*}
\tau(x) = \E_p[Y_1 - Y_0 \mid X=x].
\end{align*}
\end{definition}
Our goal is to obtain an estimate $\hat \tau(x)$ of the ITE $\tau(x)$. A natural metric is our accuracy for predicting $\tau(x)$ for a unit chosen at random from distribution $p$.
\begin{definition}
\rm
The expected \emph{precision in estimation of heterogenous effect (PEHE)}~\citep{hill} is
\begin{align*}
\epsPehe(\hat \tau) = \int_\X (\hat \tau (x) - \tau(x))^2 p(x) dx.
\end{align*}
\end{definition}
Given observational data $\DF=\{(x_i,t_i,y_{t_i,i})\}_{i=1}^n$, our goal is to estimate $\tau(x)$. One way to do so is by estimating $\hat{f}(x,t)\approx m(x,t)=\E_p[Y_t\mid x]$, and then using $\tau_{\hat{f}}(x)=\hat{f}(x,1)-\hat{f}(x,0)$. We denote $\epsPehe(\hat{f})=\epsPehe(\tau_{\hat{f}})$. Na\"{i}vely, we can use supervised learning to fit
\begin{align*}
f^0=\operatorname*{\arg\min}_{f\in\M}\mathbb{E}_p[(y_t-f(x,t))^2].
\end{align*}
Given samples $(x,y_t)$ from $p(x, y_t\mid T = t)$, we can replace the expectation in the objective with an estimate. However, when evaluating the PEHE, we are also concerned with the errors of $\hat{f}(x,t)$ on the \emph{counterfactual} distribution $p(x, y_t\mid T=1-t)$---i.e., we also need samples $(x,y_t)\sim p(x,y_t \mid T=1-t)$; otherwise, our estimate $\hat\tau(x)$ may be biased. Unfortunately, we do not have access to these kinds of samples. As we describe, our algorithm addresses this issue by using an \emph{oracle model} $f^*$ to generate data from the counterfactual distribution.

\section{Learning Causal Interpretable Models}
\label{sec:alg}

\begin{algorithm}[t]
\begin{algorithmic}
\Procedure{LearnCausal}{Factual observations $\DF=\{(x_i,t_i,y_{t_i,i})\}_{i=1}^n$, Oracle model $f^*$, Interpretable learning algorithm $\mathcal{A})$}
\State $\D_0 \gets \{(x_i, t_i)\} \cup \{(x_i, 1-t_i)\}$
\State $\D_{f^*} \gets \{(x, t, f^*(x, t)) \mid (x, t) \in \D_0 \}$
\State $\hat f \gets \A(\D_{f^*})$
\State \textbf{return} $\hat f$
\EndProcedure
\end{algorithmic}
\caption{Learning interpretable models with causal guarantees.}
\label{alg:main}
\end{algorithm}

Our learning algorithm takes three inputs: (i) interpretable learning algorithm $\mathcal{A}$ for the supervised setting., (ii) a blackbox \emph{oracle model} $f^*$ trained to predict individual treatment effects (ITEs), and (iii) an observational dataset $\DF=\{(x_i,t_i,y_i)\}_{i=1}^n$ of individuals from the factual distribution $(X,T,Y_T)$. Then, our algorithm outputs an interpretable model for predicting ITEs. More precisely, let $\M \subseteq \{f: \X \times \T \to \Y\}$ be the space of interpretable models learned by $\A$. Our goal is to learn an interpretable model $\hat{f}\in\M$ for which we can provide causal guarantees. At a high level, our algorithm uses $\mathcal{A}$ to train $\hat{f}:\X\times\T\to\Y$ to approximate $f^*$. Intuitively, if $\epsPehe(f^*)$ is small, then this approach should ensure that $\epsPehe(\hat{f})$ is small as well.

We begin by formalizing the interpretable supervised learning algorithm $\mathcal{A}$. Let $\DD = \bigcup_{n=1}^\infty \prod_{i=1}^n (\X \times \T \times \Y)$ be the set of datasets of any finite size (i.e., of size $n$ for $n\in\mathbb{N}$). Suppose we have a learning algorithm $\A:\DD\to\M$ for interpretable models---i.e., given a dataset $\D=\{(x_i,t_i,y_i)\}_{i=1}^n\in\DD$, then $\A$ (usually approximately) solves the supervised learning problem
\begin{align}
\label{eqn:aprob}
\A(\D)=\operatorname*{\arg\min}_{f\in\M}\sum_{i=1}^n(f(x_i,t_i)-y_i)^2.
\end{align}
Now, given $\mathcal{A}$, $f^*$, and some set $\D_0=\{(x_i,t_i)\}_{i=1}^{n'}$ of covariate-treatment pairs to be specified later, our algorithm compresses $f^*$ into an interpretable model $\hat f$ by constructing the training dataset
\begin{align*}
\D_{f^*}=\{(x_i,t_i,f^*(x_i,t_i)))\}_{i=1}^{n'}
\end{align*}
and then using $\A$ on $\D_{f^*}$---i.e., $\hat f=\A(\D_{f^*})$. The key question is how to choose $\D_0$ so that $\hat f$ produces a good estimate of $\tau(x)$---i.e., $\epsPehe$ is small. Intuitively, when we have control over the treatment assignment---e.g., in a randomized controlled trial (RCT)---a good distribution to use is to uniformly randomly assign treatments. In particular, consider the following distribution:
\begin{definition}
\label{def:qp}
\rm
Given distribution $p(x)$ on $\X$, the \emph{RCT distribution} $q_p(x,t)$ over $\X \times \T$ is
\begin{align*}
\P_{q_p}[T=0]&=\P_{q_p}[T=1]=1/2 \\
q_p(x\mid T=0)&=q_p(x\mid T=1)=p(x).
\end{align*}
\end{definition}
In other words, the random variables $(X,T)$ have joint distribution $q_p$ if $X\sim p(x)$, $T\sim\text{Bernoulli}(1/2)$, and $X$ and $T$ are independent. Letting $p(x)$ be the empirical distribution over covariates $x\in\X$, then $q_p$ is a good choice for $\D_0$. In particular, our algorithm (summarized in Algorithm~\ref{alg:main}) uses the distribution $\D_0=q_p$, where $p$ is the empirical distribution of covariates in $\DF$. Next, our algorithm uses $f^*$ to label the points in $\D_0$, producing a dataset $\D_{f^*}$; this step amounts to using $f^*$ to label the unobserved counterfactual for each covariate $x_i$ in $\DF$. Finally, our algorithm runs the interpretable learning algorithm $\A$ on the training set $\D_{f^*}$, and returns the result $\hat f=\A(\D_{f^*})$. As we show in Section~\ref{sec:theory}, with the choice $\D_0=q_p$, we can prove a bound on $\epsPehe$.

\section{Theoretical Guarantees}
\label{sec:theory}

In this section, we provide two theoretical guarantees for our algorithm. First, we prove that if the interpretable model $\hat f$ is a good approximation of the oracle $f^*$, then the error of $\hat f$ is also small. However, in general we may expect the gap between $\hat f$ and $f^*$ to be large. Second, we prove that under standard assumptions about the interpretable model family and the algorithm $\mathcal{A}$, we can in fact bound the error of $\hat f$ with respect to the ``best possible'' interpretable model $\tilde f$.

Finally, we discuss how our second result can be used to understand the benefits of using an indirect approach, where we train an interpretable model to mimic the oracle, compared to using an approach that directly learns an interpretable model. In particular, under reasonable assumptions, we show that the cost of using the indirect approach can be small compared to the potential gain.

\textbf{RCT error.}
We show a general connection between $\epsPehe$ and the error on the RCT distribution $q_p$.
\begin{definition}
\rm
Given a model $f:\X\times\T\to\Y$, the \emph{RCT error} of $f$ is
\begin{align*}
\epsRct(f)
&=\mathbb{E}_{q_p(x,t)}[(f(x,t)-m(x,t))^2] \\
&=\int_{\X\times\T}(f(x,t)-f^*(x,t))^2q_p(x,t)dxdt.
\end{align*}
\end{definition}
This quantity is the mean squared error (MSE) of $f$ on the RCT distribution $q_p$---i.e., it is the supervised learning loss we would use to train $f$ if we had data from the RCT distribution $q_p(x,t)$.
\begin{lemma}
\label{lem:key}
For any $f:\X\times\T\to\Y$, we have
\begin{align*}
\frac{1}{4}\epsPehe(f)&\le\epsRct(f).
\end{align*}
\end{lemma}
We give a proof in Appendix~\ref{sec:lemkeyproof}.

\textbf{Relative error bound.}
We prove that as long as $\hat{f}\in\M$ is close to $f^*$ on the distribution $q_p(x,t)$, where $p$ is the true covariate distribution, then $\epsPehe(\hat f)$ is small.
\begin{definition} \label{def:epsFf}
\rm
The \emph{relative error} of $f$ to $f^*$ is
\begin{align*}
\epsFf=\mathbb{E}_{q_p}[ (f(x,t) - f^*(x,t))^2].
\end{align*}
\end{definition}
In other words, $\epsFf$ captures the error of $f$ relative to the oracle model $f^*$.
\begin{lemma}
\label{lem:rel}
For any function $f:\X\times\T\to\Y$, and any function $f^*:\X\times\T\to\Y$, we have
\begin{align*}
\frac{1}{8}\epsPehe(f)\le\epsilon(f,f^*)+\epsRct(f^*).
\end{align*}
\end{lemma}
We give a proof in Appendix~\ref{sec:relproof}. This bound has two terms: (i) $\epsilon(f,f^*)$ captures how well $f$ approximates $f^*$, and (ii) $\epsRct(f^*)$ captures the error of the oracle $f^*$. While this bound is stated in terms of exact errors, it easily extends to a finite sample bound using standard assumptions---e.g., that the family $\M$ has finite Rademacher complexity~\citep{bartlett2002rademacher} and that $\A$ solves (\ref{eqn:aprob}) exactly.

\textbf{Optimal interpretable model bound.}
We now show how to bound the error compared to the ``best possible'' model in the model family. In particular, let
\begin{align*}
\tilde{f}&=\operatorname*{\arg\min}_{f\in \M}\epsRct(f) \\
f^0&=\operatorname*{\arg\min}_{f\in\M}\epsilon(f,f^*)
\end{align*}
be the best interpretable model for the RCT error, and the interpretable model that best approximates $f^*$ given infinite data, respectively.
\begin{lemma}
\label{lem:abs}
We have
\begin{align*}
\frac{1}{16}\epsPehe(f^0)&\le2\epsRct(f^*)+\epsRct(\tilde{f}).
\end{align*}
\end{lemma}
We give a proof in Appendix~\ref{sec:absproof}. Next, we extend Lemma~\ref{lem:abs} to account for generalization error. We assume the interpretable learning algorithm $\A$ finds the global optimizer of the empirical loss:
\begin{assumption}
\rm
The algorithm $\mathcal{A}$ solves (\ref{eqn:aprob}) exactly.
\end{assumption}
\begin{theorem}
\label{thm:gen}
We have
\begin{align*}
\frac{1}{16}\epsPehe(\hat{f})\le2\epsRct(f^*)+\epsRct(\tilde{f})+\frac{1}{2}G(n')
\end{align*}
where
\begin{align*}
G(n')&=4\Ra_{n'}(\C)+\sqrt{\frac{2\log(2/\delta)}{n'}},
\end{align*}
and where $\C=\{(x,t)\mapsto(f(x,t)-f^*(x,t))^2\mid f\in\M\}$ is the loss class, $n'=2|\DF|$ is the training set size, and $\Ra_{n'}(\C)$ is the empirical Rademacher complexity of $\C$.
\end{theorem}
We give a proof in Appendix~\ref{sec:genproof}.

\textbf{Discussion.}
The bound in Theorem~\ref{thm:gen} has three terms: (i) the error $2\epsRct(f^*)$ of the oracle $f^*$ on the RCT distribution $q_p$, (ii) the error $\epsRct(\tilde{f})$ of the best possible interpretable model on the RCT distribution, and (iii) the generalization error $G(n')/2$. In contrast, if we had access to data $D_q=\{(x_i,y_{t_i,i},t_i)\}_{i=1}^{n'}$ from the RCT distribution $(x_i,y_{t_i,t_i},t_i)\sim q_p$, a natural approach would be to use $\hat{f}'=\mathcal{A}(D_q)$. By Lemma~\ref{lem:key} and standard generalization bounds, we have
\begin{align*}
\frac{1}{4}\epsPehe(\hat{f}')\le\epsRct(\hat{f}')\le\epsRct(\tilde{f})+G(n').
\end{align*}
Our bound differs in terms of (i) the extra term $\epsRct(f^*)$, and (ii) a constant multiplicative factor. In other words, these two differences capture the ``price'' of not having access to the RCT distribution.

These results validate our hypothesis that if there are good state-of-the-art algorithms for learning blackbox models $f^*$ for causal inference, we can correspondingly obtain good algorithms for learning interpretable models $\hat{f}$ for causal inference. In particular, assuming the blackbox model is at least as good as the best interpretable model---i.e., $\epsRct(f^*)\le\epsRct(\tilde{f})$---then this approach is optimal up to a constant multiplicative factor.

\section{Causal Representations}
\label{sec:causal}

While our framework can be used with any oracle model $f^*$, using causal representations~\citep{johansson2016learning,shalit2017} to learn $f^*$ allows us to obtain end-to-end theoretical guarantees for $\hat{f}$. Recall that the key challenge in causal inference is that we do not have access to samples from the counterfactual distribution $p(x,y_t\mid T=1-t)$. If we directly fit an oracle model $f^*$ on samples $(x,t,y_t)$ from the factual distribution $p(x,y_t\mid T=t)$, then our estimator may perform poorly on the counterfactual distribution and therefore may be biased. In this case, $\epsPehe$ contains a term that comes from the discrepancy between the factual and counterfactual distributions. First, we make the following standard assumption~\citep{johansson2016learning,shalit2017}.
\begin{assumption}
\rm
The treatment assignment is \emph{strongly ignorable}---i.e.,
\begin{align*}
(Y_1,Y_0) \independent T \mid X.
\end{align*}
Furthermore, for all $x\in\X$,
\begin{align*}
0 < \P_p(T=1\mid X=x) < 1.
\end{align*}
\end{assumption}
For example, the first part eliminates the possibility that we only observe $Y_1$ for which $Y_1 > Y_0$, and the second eliminates the possibility that we never get observations of $Y_1$ for a particular $x$. Then, the factual distribution is
\begin{align*}
&p(x,y_t\mid T=t) \\
&=p(y_t\mid X=x,T=t) \cdot p(x\mid T=t) \\
&=p(y_t\mid X=x,T=1-t)\cdot p(x \mid T=t),
\end{align*}
where the last step follows by strong ignorability. In comparison, the counterfactual distribution is
\begin{align*}
&p(x,y_t\mid T=1-t) \\
&=p(y_t\mid X=x,T=1-t)\cdot p(x\mid T=1-t).
\end{align*}
The difference between these factual and counterfactual distributions is captured by the term $p(x \mid T=0)$ in the factual distribution and the term $p(x \mid T=1)$ in the counterfactual distribution.
\begin{definition}
\rm
The \emph{distribution of control units} is $p^0(x)$, and the \emph{distribution of treated units} is $p^1(x)$, where $p^t(x)=p(x\mid T=t)$.
\end{definition}
For this source of error to be small, we need $p^0(x)$ to be similar to $p^1(x)$. However, for observational data, unlike RCT data, these distributions are given to us, and are not ones that we can choose.

We consider an oracle based on causal representations~\citep{johansson2016learning,shalit2017}, which has two steps. First, learn an embedding $\Phi:\X\to\cR$, where $\cR\subseteq\R^{\ell}$, that aims to equalize the distributions $p_{\Phi}^0(r)$ and $p_{\Phi}^1(r)$ over $\cR$ induced by $\Phi$. Intuitively, if these distributions are similar, then the error term in $\epsPehe$ due to the discrepancy between $p^0(x)$ and $p^1(x)$ is small. Second, use supervised learning to train a model $h^*:\cR\times\T\to\Y$ on the dataset $\{(\Phi(x),t,y_t)\mid(x,t,y_t)\in\DF\}$, and let $f^*(x,t)=h^*(\Phi(x),t)$. They prove a bound on the error $\epsPehe$ that has two terms. The first term captures the generalization error of training $h^*$---i.e., the error of $f^*$ on the factual distribution:
\begin{definition}
\rm
The \emph{expected factual loss} of $f:\X\times\T\to\Y$ is
\begin{align*}
\epsF(f)=\E_{p(x,t)}[(m(x,t)-f(x,t))^2].
\end{align*}
\end{definition}
The second term measures the discrepancy between $p_{\Phi}^0(r)$ and $p_{\Phi}^1(r)$ using the following metric:
\begin{definition}
\label{def:IPM}
\rm
Suppose we have two probability distributions $p$ and $q$ on $\mathcal{S}\subseteq\R^d$. Given a family of functions $G\subseteq\{g:\mathcal{S}\to\R\}$, the \emph{integral probability metric (IPM)} of $p$ and $q$ is
\begin{align*}
\IPM_G (p,q)=\sup_{g\in G}\left|\int_{\mathcal S}g(s)(p(s)-q(s))ds\right|
\end{align*}
\end{definition}
\begin{assumption}
\label{assump:g}
\rm
$\Phi$ is twice-differentiable and bijective. For some $B_\Phi>0$, the family $G\subseteq\{g:\cR\to\R\}$ satisfies $B_\Phi^{-1}\cdot\ell_{h,\Phi}(\Phi^{-1}(r),t)\in G$ for each $t\in\T$, where $\ell_f(x,t)=(f(x,t)-m(x,t))^2$.
\end{assumption}
This assumption differs slightly from the one in~\citep{shalit2017}; in particular, we have stated the loss of $f(x,t)$ with respect to the expectation $m(x,t)$ rather than the ground truth $y_t$. This modification enables us to state our main result in terms of the factual distribution $\epsF$ alone. Then, we have~\citep{shalit2017}:
\begin{theorem}
\label{thm:cr}
For any $f:\X\times\T\to\Y$ of form
\begin{align*}
f(x,t)=h(\Phi(x),t)
\end{align*}
for some $h:\cR\times\T\to\Y$,
\begin{align*}
2\epsRct(f)\le p_{\text{min}}^{-1}\cdot\epsF(f)+B_\Phi\cdot\IPM_G(p_\Phi^0,p_\Phi^1),
\end{align*}
where
\begin{align*}
p_{\text{min}}=\min\{\P_p(T=0),\P_p(T=1)\}.
\end{align*}
\end{theorem}
This theorem is similar to Theorem 1 in~\citep{shalit2017}, with two modifications: (i) we have started from the RCT error $\epsRct(f)$, which is required our theoretical guarantees in Section~\ref{sec:theory}, and (ii) we have incorporated the three terms $\epsF^{t=0}(f)$, $\epsF^{t=1}(f)$, and $\sigma_Y(p)$ in their bound into the single term $p_{\text{min}}^{-1}\cdot\epsF(f)$. The second modification follows using their proof strategy with our modified version of Assumption~\ref{assump:g}. We give a proof in Appendix~\ref{sec:thmcrproof}.
\begin{corollary}
\label{cor:gencr}
We have
\begin{align*}
\frac{1}{16}\epsPehe(\hat{f})\le&p_{\text{min}}^{-1}\cdot\epsF(f^*)+B_\Phi\cdot\IPM_G(p_\Phi^0,p_\Phi^1) \\
&\qquad+\epsRct(\tilde{f})+\frac{1}{2}G(n').
\end{align*}
\end{corollary}
This result follows immediately from Theorems~\ref{thm:gen} \&~\ref{thm:cr}.

\begin{table*}[t]
\centering
\begin{tabular}{lrrrr}
\toprule
\multirow{2}{*}{{\bf Model}} & \multicolumn{2}{c}{$\sqrt{\epsPehe}$} & \multicolumn{2}{c}{$\epsilon_{\text{ATE}}$} \\
& \multicolumn{1}{c}{Ours} & \multicolumn{1}{c}{Baseline} & \multicolumn{1}{c}{Ours} & \multicolumn{1}{c}{Baseline} \\
\midrule
CFR-Net               & \multicolumn{1}{c}{--} & 0.926 $\pm$ 0.02 & \multicolumn{1}{c}{--} & 0.271 $\pm$ 0.01 \\
CART (depth 6)        & {\bf 3.668} $\pm$ 0.17 & 4.305 $\pm$ 0.20 & {\bf 0.485} $\pm$ 0.03 & 0.679 $\pm$ 0.04 \\
CART (depth 5)        & {\bf 3.824} $\pm$ 0.18 & 4.436 $\pm$ 0.21 & {\bf 0.492} $\pm$ 0.02 & 0.725 $\pm$ 0.05 \\
CART (depth 4)        & {\bf 4.086} $\pm$ 0.19 & 4.605 $\pm$ 0.22 & {\bf 0.530} $\pm$ 0.03 & 0.717 $\pm$ 0.05 \\
CART (depth 3)        & {\bf 4.462} $\pm$ 0.21 & 4.930 $\pm$ 0.23 & {\bf 0.585} $\pm$ 0.03 & 0.795 $\pm$ 0.05 \\
Honest Tree (depth 6) & {\bf 3.694} $\pm$ 0.17 & 4.086 $\pm$ 0.19 & {\bf 0.481} $\pm$ 0.02 & 0.483 $\pm$ 0.03 \\
Honest Tree (depth 5) & {\bf 3.760} $\pm$ 0.17 & 4.098 $\pm$ 0.19 & 0.488 $\pm$ 0.02 & {\bf 0.486} $\pm$ 0.03 \\
Honest Tree (depth 4) & {\bf 3.875} $\pm$ 0.18 & 4.128 $\pm$ 0.19 & 0.498 $\pm$ 0.02 & {\bf 0.488} $\pm$ 0.03 \\
Honest Tree (depth 3) & {\bf 4.090} $\pm$ 0.19 & 4.237 $\pm$ 0.20 & 0.535 $\pm$ 0.03 & {\bf 0.498} $\pm$ 0.03 \\
LASSO                 & {\bf 5.725} $\pm$ 0.26 & 5.777 $\pm$ 0.26 & {\bf 0.671} $\pm$ 0.04 & 0.942 $\pm$ 0.05 \\
Kernel Ridge          & {\bf 2.077} $\pm$ 0.09 & 3.190 $\pm$ 0.14 & {\bf 0.361} $\pm$ 0.02 & 0.562 $\pm$ 0.02 \\
GBM                   & {\bf 1.845} $\pm$ 0.09 & 2.799 $\pm$ 0.14 & {\bf 0.352} $\pm$ 0.02 & 0.453 $\pm$ 0.03 \\
Random Forest         & {\bf 2.905} $\pm$ 0.14 & 3.653 $\pm$ 0.19 & {\bf 0.439} $\pm$ 0.02 & 0.621 $\pm$ 0.04 \\
\bottomrule
\end{tabular}
\caption{We show results comparing our approach to a baseline estimator for a number of model families on the IHDP dataset. For each value, we show the mean $\pm$ the standard error. We bold the better of the two values between ours and the baseline.}
\label{tab:fullresults}
\end{table*}

\section{Experiments}
\label{sec:exp}

\begin{figure*}
\centering
\begin{tabular}{cc}
\includegraphics[width=0.4\textwidth]{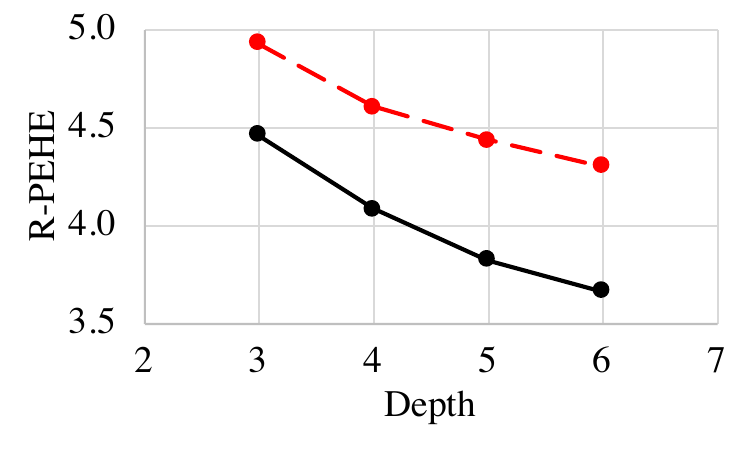} \hspace{0.1in} & \hspace{0.1in}
\includegraphics[width=0.4\textwidth]{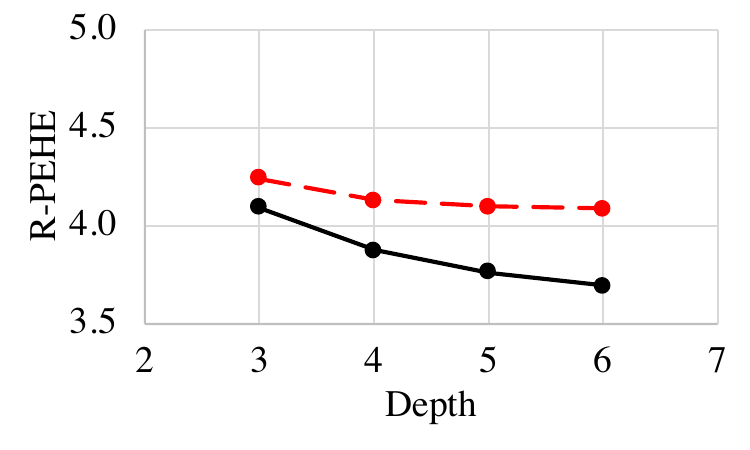}
\end{tabular}
\caption{Performance (in terms of $\sqrt{\epsPehe}$) of CART (left) and honest trees (right) using our approach (black, solid) and the baseline approach (red, dashed), as a function of the depth of the decision tree.}
\label{fig:cartdepth}
\end{figure*}

As discussed previously, our focus is on showing that the models we train can improve performance for a fixed supervised learning algorithm. For any interpretable supervised learning algorithm chosen by the user, they can use this algorithm within our framework to convert that algorithm to one for predicting ITEs. Evaluating the performance of causal models is a challenging task, since ground truth data on individual treatment effects (ITEs) is difficult to obtain. Following previous work~\citep{shalit2017}, we evaluate our framework on the IHDP \citep{hill} dataset.

\textbf{Dataset.}
We use a dataset for causal inference evaluation based on the Infant Health and Development Program, from \citep{hill} and preprocessed by \citep{shalit2017} using the NPCI package \citep{NPCI}. The dataset has 747 units (139 treated, 708 control) and 25 covariates of children and their mothers. This dataset contains 1000 realizations of the outcomes with 63/27/10 train/validation/test splits. The outcomes in this dataset are simulated, so we have ground truth values of the ITE for each unit. Using this ground truth, we can obtain a test set estimate $\hatepsPehe(f)$ of the error in the predicted ITE. Then, we report the mean and standard errors of $\sqrt{\hatepsPehe(f)}$, as well as the absolute error in the ATE
\begin{align*}
\epsilon_{\text{ATE}}
=&~\left|\frac{1}{n}\sum_{i=1}^n(\hat{\tau}(x_i)-\tau(x_i))\right|
\end{align*}
over the 1000 realizations. Our primary metric of interest is $\sqrt{\hatepsPehe(f)}$, since it measures predictive accuracy of ITEs; in contrast, $\epsilon_{\text{ATE}}$ measures predictive accuracy of ATEs.

\textbf{Oracle model.}
For $f^*$, we train a CFR-net from \citep{shalit2017}, which has 3 fully connected exponential-linear layers for each the embedding $\Phi$ and for the prediction function $h^*$, with layer sizes 200 and 100 for the representation and hypothesis layers. We use mean squared error.

\textbf{Interpretable models.}
We evaluate the performance of our approach on a variety of models with a range of interpretability: CART trees~\citep{breiman2017classification}, honest trees~\citep{athey2016recursive}, LASSO regression~\citep{tibshirani1996regression}, kernel ridge regression~\citep{murphy2012machine}, gradient boosted models (GBMs)~\citep{friedman2001greedy}, and random forests~\citep{breiman2001random}. For each model family, we train one model using our approach, and a baseline model using only the observational data. Of these models, only honest trees are designed to handle causality; however, their focus is on obtaining unbiased estimates rather than low-variance estimates. In particular, they split the dataset into two, using the first part to estimate splits and the second to estimate values at the leaf nodes. This approach ensures that the estimates at the leaf nodes are unbiased, but also greatly increases variance since they are only using half the data at each point.

\textbf{Results.}
We show results in Table~\ref{tab:fullresults}. Also, we run CART and honest trees with different maximum depths; Figure~\ref{fig:cartdepth} shows how $\sqrt{\epsPehe}$ scales with depth.

\textbf{Discussion.}
Our approach uniformly outperforms the baseline approach in terms of $\sqrt{\epsPehe}$, which measures performance on predicting ITEs. Even on predicting ATEs, our approach mostly outperforms the baseline; the only exception are honest trees, which are interpretable models tailored towards estimating treatment effects. As we discussed before, honest trees are focused on reducing bias at the expense of increased variance. Otherwise, we observe the usual trends---more complex models (e.g., GBMs and random forests) outperform more interpretable models (LASSO, CART, honest trees). In summary, our results clearly demonstrate the potential for our approach to substantially improve the performance of interpretable learning algorithms used to predict ITEs.

\section{Conclusion}
\label{sec:conc}

We have proposed a general framework for learning interpretable models with causal guarantees. A key direction for future work is designing oracle models that do not rely on strong ignorability to obtain provable guarantees.


\bibliography{paper}
\bibliographystyle{plainnat}

\clearpage

\onecolumn
\appendix
\section{Proofs}

\subsection{Proof of Lemma~\ref{lem:key}}
\label{sec:lemkeyproof}

We have
\begin{align*}
\epsPehe(f)
=&\int_\X((f(x,1)-f(x,0))-(m(x,1)-m(x,0)))^2p(x)dxdt \\
=&\int_\X((f(x,1)-m(x,1))-(f(x,0)-m(x,0)))^2p(x)dxdt \\
\le&2\int_\X(f(x,0)-m(x,0))^2p(x)dx+2\int_\X(f(x,1)-m(x,1))^2p(x)dx \\
=&2\int_\X(f(x,0)-m(x,0))^2p(x,0)dx+2\int_\X(f(x,0)-m(x,0))^2p(x,1)dx \\
&\qquad+2\int_\X(f(x,1)-m(x,1))^2p(x,0)dx+2\int_\X(f(x,1)-m(x,1))^2p(x,1)dx \\
=&2\int_\X(f(x,t)-m(x,t))^2p(x,t)dxdt+2\int_\X(f(x,t)-m(x,t))^2p(x,1-t)dxdt \\
=&4\int_\X\ell(f(x,t),m(x,t))q_p(x,t)dxdt,
\end{align*}
as claimed. $\qed$

\subsection{Proof of Lemma~\ref{lem:rel}}
\label{sec:relproof}

By Lemma~\ref{lem:key}, we have
\begin{align*}
\frac{1}{8}\epsPehe(f)
&\le\frac{1}{2}\mathbb{E}_{q_p(x,t)}[(f(x,t)-m(x,t))^2] \\
&=\frac{1}{2}\mathbb{E}_{q_p(x,t)}[((f(x,t)-f^*(x,t))+(f^*(x,t)-m(x,t)))^2] \\
&\le\mathbb{E}_{q_p(x,t)}[(f(x,t)-f^*(x,t))^2]+\mathbb{E}_{q_p(x,t)}[(f^*(x,t)-m(x,t))^2] \\
&=\epsilon(f,f^*)+\epsRct(f^*),
\end{align*}
as claimed. $\qed$

\subsection{Proof of Lemma~\ref{lem:abs}}
\label{sec:absproof}

Note that
\begin{align*}
\frac{1}{16}\epsPehe(f^0)-\frac{1}{2}\epsRct(f^*)
&\le\frac{1}{2}\epsilon(f^0,f^*) \\
&\le\frac{1}{2}\epsilon(\tilde{f},f^*) \\
&=\frac{1}{2}\mathbb{E}_{q_p(x,t)}[(\tilde{f}(x,t)-f^*(x,t))^2] \\
&=\frac{1}{2}\mathbb{E}_{q_p(x,t)}[((\tilde{f}(x,t)-m(x,t))-(f^*(x,t)-m(x,t)))^2] \\
&\le\mathbb{E}_{q_p(x,t)}[(\tilde{f}(x,t)-m(x,t))^2]+\mathbb{E}_{q_p(x,t)}[f^*(x,t)-m(x,t))^2] \\
&=\epsRct(\tilde{f})+\epsRct(f^*),
\end{align*}
where the first step follows by Lemma~\ref{lem:rel} and the second step follows by the definition of $f^0$. $\qed$

\subsection{Proof of Theorem~\ref{thm:gen}}
\label{sec:genproof}

For any $\delta>0$, with probability at least $1-\delta$, we have
\begin{align*}
\frac{1}{16}\epsPehe(\hat{f})-\frac{1}{2}\epsRct(f^*)
&\le\frac{1}{2}\epsilon(\hat{f},f^*) \\
&\le\frac{1}{2}\epsilon(f^0,f^*)+G(n') \\
&\le\epsRct(\tilde{f})+\epsRct(f^*)+G(n'),
\end{align*}
where the first step follows by Lemma~\ref{lem:rel}, the second step follows from generalization bounds based on Rademacher complexity~\cite{bartlett2002rademacher,liang2016statistical}, and the third step follows by the proof of Lemma~\ref{lem:abs}. $\qed$

\subsection{Proof of Theorem~\ref{thm:cr}}
\label{sec:thmcrproof}

Define
\begin{align*}
\epsF(f)&=\E_{p(x,t)}[\ell_f(x,t)] \\
\epsCF(f)&=\E_{p(x,1-t)}[\ell_f(x,t)] \\
\epsF^t(f)&=\E_{p^t(x)}[\ell_f(x,t)] \\
\epsCF^t(f)&=\E_{p^{1-t}(x)}[\ell_f(x,t)].
\end{align*}
and let $p_t=\mathbb{P}_p(T=t)$. Then, we have
\begin{align*}
\epsCF(f)-\sum_{t\in\T}p_{1-t}\cdot\epsF^t(f)
&=\sum_{t\in\T}p_{1-t}\cdot(\epsCF^t(f)-\epsF^t(f)) \\
&=\sum_{t\in\T}p_{1-t}\cdot\int_{\X}\ell_f(x,t)\cdot(p^{1-t}(x)-p^t(x))dx \\
&=\sum_{t\in\T}p_{1-t}\cdot\int_{\cR}\ell_f(\Phi^{-1}(r),t)\cdot(p_{\Phi}^{1-t}(r)-p_{\Phi}^t(r))dr \\
&\le\sum_{t\in\T}p_{1-t}\cdot B_{\Phi}\cdot\IPM_G(p_{\Phi}^{1-t},p_{\Phi}^t) \\
&=B_{\Phi}\cdot\IPM_G(p_{\Phi}^0,p_{\Phi}^1).
\end{align*}
Now, note that
\begin{align*}
2\epsRct(f)
&=2\int_{\X\times\{0,1\}}(f(x,t)-m(x,t))^2q_p(x, t)dxdt \\
&=\int_{\X\times\{0,1\}}(f(x,t)-m(x,t))^2(p(x,t)+p(x,1-t))dxdt \\
&=\epsF(f)+\epsCF(f) \\
&\le\epsF^0(f)+\epsF^1(f)+B_\Phi\cdot\IPM_G(p_\Phi^0,p_\Phi^1) \\
&\le p_{\text{min}}^{-1}(p_0\cdot\epsF^0(f)+p_1\cdot\epsF^1(f))+B_\Phi\cdot\IPM_G(p_\Phi^0,p_\Phi^1) \\
&=p_{\text{min}}^{-1}\cdot\epsF(f)+B_\Phi\cdot\IPM_G(p_\Phi^0,p_\Phi^1),
\end{align*}
as claimed. $\qed$

\end{document}